# MoNet: Moments Embedding Network[*]

Mengran Gou[1]  Fei Xiong[2]  Octavia Camps[1]  Mario Sznaier[1]
[1]Electrical and Computer Engineering, Northeastern University, Boston, MA, US
[2]Information Sciences Institute, USC, CA, US
{mengran, camps, msznaier}@coe.neu.edu
feixiong@ads.isi.edu

## Abstract

*Bilinear pooling has been recently proposed as a feature encoding layer, which can be used after the convolutional layers of a deep network, to improve performance in multiple vision tasks. Different from conventional global average pooling or fully connected layer, bilinear pooling gathers 2nd order information in a translation invariant fashion. However, a serious drawback of this family of pooling layers is their dimensionality explosion. Approximate pooling methods with compact properties have been explored towards resolving this weakness. Additionally, recent results have shown that significant performance gains can be achieved by adding 1st order information and applying matrix normalization to regularize unstable higher order information. However, combining compact pooling with matrix normalization and other order information has not been explored until now. In this paper, we unify bilinear pooling and the global Gaussian embedding layers through the empirical moment matrix. In addition, we propose a novel sub-matrix square-root layer, which can be used to normalize the output of the convolution layer directly and mitigate the dimensionality problem with off-the-shelf compact pooling methods. Our experiments on three widely used fine-grained classification datasets illustrate that our proposed architecture, MoNet, can achieve similar or better performance than with the state-of-art $G^2$DeNet. Furthermore, when combined with compact pooling technique, MoNet obtains comparable performance with encoded features with 96% less dimensions.*

## 1. Introduction

Embedding local representations of an image to form a feature that is representative yet invariant to nuisance noise is a key step in many computer vision tasks. Before the phenomenal success of deep convolutional neural networks (CNN) [18], researchers tackled this problem with hand-crafted consecutive independent steps. Remarkable works include HOG [6], SIFT [24], covariance descriptor [33], VLAD [14], Fisher vector [27] and bilinear pooling [3]. Although CNNs are trained from end to end, they can be also viewed as two parts, where the convolutional layers are feature extraction steps and the later fully connected (FC) layers are an encoding step. Several works have been done to explore substituting the FC layers with conventional embedding methods in both two-stage fashion [4, 11] and end-to-end trainable way [22, 13].

Bilinear CNN (BCNN) was first proposed by Lin *et al.* [22] to pool the second order statistics information across the spatial locations. Bilinear pooling has been proven to be successful in many tasks, including fine-grained image classification [16, 9], large-scale image recognition [20], segmentation [13], visual question answering [8, 37], face recognition [29] and artistic style reconstruction [10]. Wang *et al.* [36] proposed to also include the 1st order informa-tion by using a Gaussian embedding in G$^2$DeNet. It has been shown that the normalization method is also critical to these CNNs performance. Two normalization methods have been proposed for the bilinear pooled matrix, $\mathbf{M} = \frac{1}{n}\mathbf{X}^T\mathbf{X}$, where $\mathbf{X} \in \mathbb{R}^{n \times C}$ represents the local features. On one hand, because $\mathbf{M}$ is Symmetric Positive Definite (SPD), Ionescu et al. [13] proposed to apply matrix-logarithm to map the SPD matrices from the Riemannian manifold to an Euclidean space, followed by $\log(\mathbf{M}) = \mathbf{U}_M \log(\mathbf{S}_M) \mathbf{U}_M^T$ with $\mathbf{M} = \mathbf{U}_M \mathbf{S}_M \mathbf{U}_M^T$. On the other hand, [36, 21] proposed matrix-power to scale $\mathbf{M}$ non-linearly with $\mathbf{M}^p = \mathbf{U}_M \mathbf{S}_M^p \mathbf{U}_M^T$. In both works, matrix-power was shown to have better performance and numerically stability than the matrix-logarithm. In addition, Li et al. [20] provided theoretical support on the superior performance of matrix-power normalization in solving a general large-scale image recognition problem.

Table 1. Comparison of 2nd order statistical information-based neural networks. Bilinear CNN (BCNN) only has 2nd order information and does not use matrix normalization. Both improved BCNN (iBCNN) and $G^2$DeNet take advantage of matrix normalization but suffer from large dimensionality since they use the square-root of a large pooled matrix. Our proposed MoNet, with the help of a novel sub-matrix square-root layer, can normalize the local features directly and reduce the final representation dimension significantly by substituting bilinear pooling with compact pooling.

|  | 1st order moment | Matrix normalization | Compact capacity |
|---|---|---|---|
| BCNN [22, 9] | ✗ | ✗ | ✓ |
| iBCNN [21] | ✗ | ✓ | ✗ |
| $G^2$DeNet [36] | ✓ | ✓ | ✗ |
| MoNet | ✓ | ✓ | ✓ |

[*]This work was supported in part by NSF grants IIS-1318145, ECCS-1404163, CMMI-1638234 and CNS-1646121; AFOSR grant FA9550-15-1-0392; and the Alert DHS Center of Excellence under Award Number 2013-ST-061-ED0001.

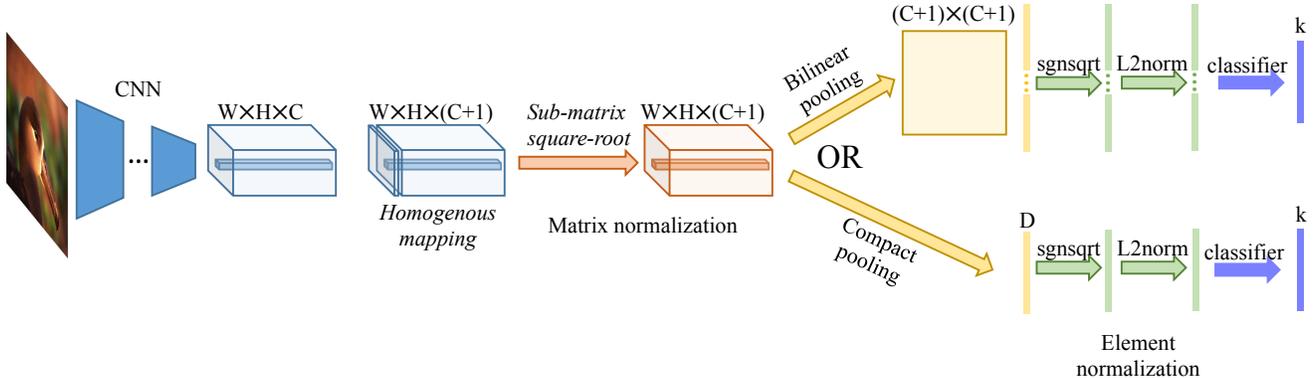

Figure 1. Architecture of the proposed moments-based network **MoNet**. With the proposed sub-matrix square-root layer, it is possible to perform matrix normalization before bilinear pooling or further apply compact pooling to reduce the dimensionality dramatically without undermining performance.

A critical weakness of the above feature encoding is the extremely high dimensionality of the encoded features. Due to the tensor product[1], the final feature dimension is $C^2$ where $C$ is the number of feature channels of the last convolution layer. Even for relatively low $C = 512$ as in VGG-16 [30], the dimensionality of the final feature is already more than $262K$. This problem can be alleviated by using random projections [9], tensor sketching [9, 5], and the low rank property [16]. However, because the matrix-power normalization layer is applied on the pooled matrix $\mathbf{M}$, it is non-trivial to combine matrix normalization and compact pooling to achieve better performance and reduce the final feature dimensions at the same time.

In this paper, we propose a new architecture, MoNet, that integrates matrix-power normalization with Gaussian embedding. To this effect, we re-write the formulation of G$^2$DeNet using the tensor product of the homogeneous padded local features to align it with the architecture of BCNN so that the Gaussian embedding operation and bilinear pooling are decoupled. Instead of working on the bilinear pooled matrix $\mathbf{M}$, we derive the sub-matrix square-root layer to perform the matrix-power normalization directly on the (in-)homogeneous local features. With the help of this novel layer, we can take advantage of compact pooling to approximate the tensor product, but with much fewer dimensions.

The main contributions of this work are three-fold:

- We unify the G$^2$DeNet and bilinear pooling CNN using the empirical moment matrix and decouple the Gaussian embedding from bilinear pooling.

- We propose a new sub-matrix square-root layer to directly normalize the features before the bilinear pooling layer, which makes it possible to reduce the dimensionality of the representation using compact pooling.

- We derive the gradient of the proposed layer using matrix back propagation, so that the whole proposed *moments embedding network* **"MoNet"** architecture can be optimized jointly.

## 2. Related work

Bilinear pooling was proposed by Tenenbaum et al. [32] to model two-factor structure in images to separate style from content. Lin et al. [22] introduced it into a convolutional neural network as a pooling layer and improved it further by adding matrix power normalization in their recent work [21]. Wang et al. [36] proposed G$^2$DeNet with Gaussian embedding, followed by matrix normalization to incorporate 1st order moment information and achieved the state-of-the-art performance. In a parallel research track, low dimension compact approximations of bilinear pooling have been also explored. Gao *et al*. [9] bridged bilinear pooling with a linear classifier with a second order polynomial kernel by adopting the off-the-shelf kernel approximation methods Random MacLaurin [15] and Tensor Sketch [28] to pool the local features in a compact way. Cui [5] generalized this approach to higher order polynomials with Tensor Sketch. By combining with bilinear SVM, Kong *et al*. [16] proposed to impose a low-rank constraint to reduce the number of parameters. However, none of these approaches can be easily integrated with matrix normalization because of the absence of a bilinear pooled matrix.

---
[1] We show that the Gaussian embedding can be written as a tensor product in sec. 3.2.1 In the following sections, we will use tensor product and bilinear pooling interchangeably.

Lasserre *et al*. [19] proposed to use the empirical moment matrix formed by explicit in-homogeneous polynomial kernel basis for outlier detection. Sznaier *et al*. [31] improved the performance for the case of data subspaces, by working on the singular values directly. In [12], the empirical moments matrix was applied as a feature embedding method for the person re-identification problem and it was shown that the Gaussian embedding [23] is a special case when the moment matrix order equals to 1. However, both of these works focus on a conventional pipeline and did not bring moments to modern CNN architectures.

Ionescu *et al*. [13] introduced the theory and practice of matrix back-propagation for training CNNs, which enable structured matrix operations in deep neural networks training. Both [21] and [36] used it to derive the back-propagation of the matrix square-root and matrix logarithm for a symmetric matrix. Li *et al*. [20] applied a generalized $p$-th order matrix power normalization instead of the square-root. However, in our case, since we want to apply the matrix normalization directly on a non-square local feature matrix, we cannot plug-in the equation directly from previous works.

## 3. MoNet Architecture

The overview of the proposed MoNet architecture is shown in Fig. 1. For an input image $\mathbf{I}$, the output of the last convolution layer after the ReLU, $\mathbf{X}$, consists of local features $\mathbf{x}_i$, across spatial locations $i = 1, 2, \ldots, n$. Then, we introduce a homogeneous mapping (HM) layer to disentangle the tensor product operator. After that, a novel sub-matrix square-root (Ssqrt) layer is applied to directly normalize the feature vector before the tensor product. Finally, a compact bilinear pooling layer pools all $n$ features across all spatial locations, followed by an element-wise square-root regularization and $\ell_2$ normalization before the final fully-connected layer. Next, we will detail the design of each block.

### 3.1. Homogeneous mapping layer

Since the global Gaussian embedding layer used in G$^2$DeNet entangles the tensor product operator, one cannot directly incorporate compact bilinear pooling. With the help of the proposed HM layer, we can re-write the Gaussian embedding layer with a HM layer followed by a tensor product, as explained next.

Assume $\mathbf{X} \in \mathbb{R}^{n \times C}$, corresponding to $n$ features with dimension $C$ and $n > C$, mean $\mu$ and covariance $\mathbf{\Sigma}$. The homogeneous mapping of $\mathbf{X}$ is obtained by padding $\mathbf{X}$ with an extra dimension set to $1$. For the simplicity of the following layers, instead of applying the conventional bilinear pooling layer as in [22], we also divide the homogeneous feature by the square-root of the number of samples. Then, the forward equation of the homogeneous mapping layer is:

$$\tilde{\mathbf{X}} = \frac{1}{\sqrt{n}}[\mathbf{1}|\mathbf{X}] \in \mathbb{R}^{n \times (C+1)} \quad (1)$$

The tensor product of $\tilde{\mathbf{X}}$ can be written as

$$\mathbf{M} = \tilde{\mathbf{X}}^T \tilde{\mathbf{X}} = \begin{bmatrix} \mathbf{1} & \mu \\ \mu^T & \frac{1}{n}\mathbf{X}^T\mathbf{X} \end{bmatrix} \quad (2)$$

where $\mu = \frac{1}{n}\sum_1^n \mathbf{X}$. Since $\frac{1}{n}\mathbf{X}^T\mathbf{X} = \mathbf{\Sigma} + \mu^T\mu$, Eq. 2 is the Gaussian embedding method used in G$^2$DeNet [36]. One can also show that the conventional bilinear pooling layer is equal to the tensor product of the in-homogeneous feature matrix.

### 3.2. Sub-matrix square-root layer

Matrix normalization in iBCNN and G$^2$DeNet requires the computation of the singular value decomposition (SVD) of the output of the tensor product, which prevents the direct use of compact bilinear pooling. We will address this issue by incorporating a novel layer, named sub-matrix square-root (Ssqrt) layer, to perform the equivalent matrix normalization before the tensor product. This choice is supported by experimental results in [36, 21] showing that the matrix square-root normalization is better than the matrix logarithm normalization for performance and training stability.

#### 3.2.1 Forward propagation

Recall that given the SVD of a SPD matrix, $\mathbf{Q} = \mathbf{U}_Q \mathbf{S}_Q \mathbf{U}_Q^T$, the square root of $\mathbf{Q}$ is defined as

$$\mathbf{Q}^{\frac{1}{2}} = \mathbf{U}_Q \mathbf{S}_Q^{\frac{1}{2}} \mathbf{U}_Q^T \quad (3)$$

where $\mathbf{S}_Q^{\frac{1}{2}}$ is computed by taking the square root of its diagonal elements.

Consider now the SVD of $\tilde{\mathbf{X}} = \mathbf{U}\mathbf{S}\mathbf{V}^T$. Then, we have

$$\mathbf{M} = \tilde{\mathbf{X}}^T \tilde{\mathbf{X}} = \mathbf{V}\mathbf{S}^T\mathbf{U}^T\mathbf{U}\mathbf{S}\mathbf{V}^T \quad (4)$$

and since $\mathbf{U}^T\mathbf{U} = \mathbf{I}$ and $\mathbf{S}^T\mathbf{S}$ is a square matrix:

$$\mathbf{M}^{\frac{1}{2}} = \mathbf{V}(\mathbf{S}^T\mathbf{S})^{\frac{1}{2}}\mathbf{V}^T \quad (5)$$

Note that $\mathbf{S} \in \mathbb{R}^{n \times (C+1)}, n > C+1$ and hence its square root is not well defined. We introduce a helper matrix $\mathbf{A}$ to keep all non-zero singular values in $\mathbf{S}$ as follows:

$$\mathbf{S} = \mathbf{A}\tilde{\mathbf{S}}, \mathbf{A} = [\mathbf{I}_{C+1}|\mathbf{0}]^T \quad (6)$$

where $\tilde{\mathbf{S}} \in \mathbb{R}^{(C+1) \times (C+1)}$ is a square diagonal matrix and $\mathbf{I}_{C+1}$ is the $(C+1) \times (C+1)$ identity matrix. Substituting Eq. (6) in Eq. (5), we have

$$\mathbf{M}^{\frac{1}{2}} = \mathbf{V}(\tilde{\mathbf{S}}\mathbf{A}^T\mathbf{A}\tilde{\mathbf{S}})^{\frac{1}{2}}\mathbf{V}^T = \mathbf{V}\tilde{\mathbf{S}}^{\frac{1}{2}}\tilde{\mathbf{S}}^{\frac{1}{2}}\mathbf{V}^T \quad (7)$$

since $\mathbf{A}^T\mathbf{A} = \mathbf{I}_{C+1}$. To keep the same number of samples for the input and output of this layer, we finally re-write Eq. (5) in the following tensor product format:

$$\mathbf{M}^{\frac{1}{2}} = \mathbf{Y}^T\mathbf{Y} \quad (8)$$

where the output $\mathbf{Y}$ is defined as $\mathbf{Y} = \mathbf{A}\tilde{\mathbf{S}}^{\frac{1}{2}}\mathbf{V}^T$, allowing us to perform matrix normalization directly on the features $\tilde{\mathbf{X}}$.

Note that because in most modern CNNs, $n$ cannot be much greater than $C$ and the features after ReLU tend to be sparse, $\tilde{\mathbf{X}}$ is usually rank deficient. Therefore, we only use the non-zero singular values and singular vectors. Then, the forward equation of the sub-matrix square-root layer can be written as

$$\mathbf{Y} = \mathbf{A}_{:,1:e}\tilde{\mathbf{S}}^{\frac{1}{2}}_{1:e}\mathbf{V}^T_{:,1:e} \quad (9)$$

where $e$ is the index of the smallest singular value greater than $\epsilon$ [2].

### 3.2.2 Backward propagation

We will follow the matrix back propagation techniques proposed by Ionescu *et al.* [13] to derive the equation of the back propagation path for the sub-matrix square-root layer.

For a scalar loss $L = f(\mathbf{Y})$, we assume $\frac{\partial L}{\partial \mathbf{Y}}$ is available when we derive the back propagation. Let $\tilde{\mathbf{X}} = \mathbf{U}\mathbf{S}\mathbf{V}^T$ and $\mathbf{U} \in \mathbb{R}^{n \times n}$. We can form $\mathbf{U}$ using block decomposition as $\mathbf{U} = [\mathbf{U}_1|\mathbf{U}_2]$ with $\mathbf{U}_1 \in \mathbb{R}^{n \times (C+1)}$ and $\mathbf{U}_2 \in \mathbb{R}^{n \times (n-C-1)}$. The partial derivatives between a given scalar loss $L$ and $\tilde{\mathbf{X}}$ are

$$\frac{\partial L}{\partial \tilde{\mathbf{X}}} = \mathbf{D}\mathbf{V}^T + \mathbf{U}(\frac{\partial L}{\partial \mathbf{S}} - \mathbf{U}^T\mathbf{D})_{diag}\mathbf{V}^T + \\ 2\mathbf{U}\mathbf{S}(\mathbf{K}^T \circ \left(\mathbf{V}^T(\frac{\partial L}{\partial \mathbf{V}} - \mathbf{V}\mathbf{D}^T\mathbf{U}\mathbf{S})\right)_{sym}\mathbf{V}^T \quad (10)$$

where $\circ$ represents element-wise product, $(\mathbf{Q})_{sym} \doteq \frac{1}{2}(\mathbf{Q}^T + \mathbf{Q})$ and

$$\mathbf{D} = \left(\frac{\partial L}{\partial \mathbf{U}}\right)_1 \tilde{\mathbf{S}}^{-1} - \mathbf{U}_2\left(\frac{\partial L}{\partial \mathbf{U}}\right)_2^T \mathbf{U}_1\tilde{\mathbf{S}}^{-1} \quad (11)$$

[2] We will omit the subscript in the following for a concise notation

$$\mathbf{K}_{ij} = \begin{cases} \frac{1}{s_i^2 - s_j^2} & i \neq j \\ 0 & i = j \end{cases} \quad (12)$$

From Eq. 9, we can compute the variation of $\mathbf{Y}$ as

$$d\mathbf{Y} = \frac{1}{2}\mathbf{A}\tilde{\mathbf{S}}^{-\frac{1}{2}}d\tilde{\mathbf{S}}\mathbf{V}^T + \mathbf{A}\tilde{\mathbf{S}}^{\frac{1}{2}}d\mathbf{V}^T \quad (13)$$

Based on the chain rule, the total variation can be written as

$$\frac{\partial L}{\partial \mathbf{Y}} : d\mathbf{Y} = \frac{1}{2}\frac{\partial L}{\partial \mathbf{Y}} : \mathbf{A}\tilde{\mathbf{S}}^{-\frac{1}{2}}d\tilde{\mathbf{S}}\mathbf{V}^T + \frac{\partial L}{\partial \mathbf{Y}} : \mathbf{A}\tilde{\mathbf{S}}^{\frac{1}{2}}d\mathbf{V}^T \quad (14)$$

where : denotes the inner-product. After re-arrangement with the rotation properties of inner-product, we re-write the above equation as

$$\frac{\partial L}{\partial \mathbf{Y}} : d\mathbf{Y} = \frac{1}{2}\tilde{\mathbf{S}}^{-\frac{1}{2}}\mathbf{A}^T\frac{\partial L}{\partial \mathbf{Y}}\mathbf{V} : d\tilde{\mathbf{S}} + \tilde{\mathbf{S}}^{\frac{1}{2}}\mathbf{A}^T\frac{\partial L}{\partial \mathbf{Y}} : d\mathbf{V}^T \quad (15)$$

Therefore, we have

$$\frac{\partial L}{\partial \mathbf{S}} = \mathbf{A}\frac{\partial L}{\partial \tilde{\mathbf{S}}} = \frac{1}{2}\mathbf{A}\tilde{\mathbf{S}}^{-\frac{1}{2}}\mathbf{A}^T\frac{\partial L}{\partial \mathbf{Y}}\mathbf{V} \quad (16)$$

$$\frac{\partial L}{\partial \mathbf{V}} = (\frac{\partial L}{\partial \mathbf{V}^T})^T = (\frac{\partial L}{\partial \mathbf{Y}})^T\mathbf{A}\tilde{\mathbf{S}}^{\frac{1}{2}} \quad (17)$$

Finally, substituting Eq. 16 and Eq. 17 into Eq. 10 and considering $\frac{\partial L}{\partial \mathbf{U}} = 0$, we have

$$\frac{\partial L}{\partial \tilde{\mathbf{X}}} = \mathbf{U}\left(\frac{1}{2}\mathbf{A}\tilde{\mathbf{S}}^{-\frac{1}{2}}\mathbf{A}^T\frac{\partial L}{\partial \mathbf{Y}}\mathbf{V} + \\ 2\mathbf{S}\left[\mathbf{K}^T \circ \left(\mathbf{V}^T\left(\frac{\partial L}{\partial \mathbf{Y}}\right)^T\mathbf{A}\tilde{\mathbf{S}}^{\frac{1}{2}}\right)\right]_{sym}\right)\mathbf{V}^T \quad (18)$$

### 3.3. Compact pooling

Following the work in [9, 8], we adopt the Tensor Sketch (TS) method to approximate bilinear pooling due to it better performance and lower computational and memory cost. Building up on count sketch and FFT, one can generate a tensor sketch function s.t. $\langle TS_1(x), TS_2(y) \rangle \approx \langle x, y \rangle^2$, using Algorithm 1. The back-propagation of a TS layer is given by [9].

As shown in Table 2, with the techniques mentioned above, the proposed MoNet is capable to solve the problem with much less computation and memory complexity than the other BCNN based algorithms.

## 4. Experiments

Aligned with other bilinear CNN based papers, we also evaluate the proposed MoNet with three widely used fine-grained classification datasets. The experimental setups and the algorithm implementation are described in detail in Sec. 4.1. Then, in Sec. 4.2, the experimental results on fine-grained classification are presented and analyzed.

Table 2. Dimension, computation and memory information for the networks we compared in this paper. $H, W$ and $C$ represent the height, width and number of feature channels for the output of the final convolution layer, respectively. $k$ and $D$ denote the number of classes and projected dimensions for Tensor Sketch, respectively. Numbers inside brackets indicate the typical value when the corresponding network was evaluated with VGG-16 model [30] on a classification task with 1,000 classes. In this case, $H = W = 13, C = 512, k = 1,000, D = 10,000$ and all data was stored with single precision.

|  | BCNN [22] | iBCNN [21] | iBCNN TS | $G^2$DeNet [36] | MoNet | MoNet TS |
|---|---|---|---|---|---|---|
| Dimension | $C^2$ [262K] | $C^2$ [262K] | $D$ [10K] | $(C+1)^2$ [263K] | $(C+1)^2$ [263k] | $D$ [10k] |
| Parameter Memory | 0 | 0 | 2C | 0 | 0 | 2C |
| Computation | $O(HWC^2)$ | $O(HWC^2)$ | $O(HW(C+D\log D))$ | $O(HWC^2)$ | $O(HWC^2)$ | $O(HW(C+D\log D))$ |
| Classifier Memory | $kC^2$ [1000MB] | $kC^2$ [1000MB] | $kD$ [40MB] | $k(C+1)^2$ [1004MB] | $k(C+1)^2$ [1004MB] | $kD$ [40MB] |

**Algorithm 1** Tensor Sketch approximation pooling

**Require:** $x$, projected dimension $D$

1: Generate randomly selected (but fixed) two pairs of hash functions $h_t \in \mathbb{R}^D$ and $s_t \in \mathbb{R}^D$ where $t = 1, 2$ and $h_t(i), s_t(i)$ are uniformly drawn from $\{1, 2, \cdots, D\}$ and $\{-1, +1\}$, respectively.
2: Define count sketch function $\Psi(x, h_t, s_t) = [\psi_1(x), \psi_2(x), \cdots, \psi_D(x)]^T$ where $\psi_j(x) = \sum_{i:h_t(i)=j} s_t(i) x_i$
3: Define $TS(x) = FFT^{-1}(FFT(\Psi(x, h_1, s_1) \circ (\Psi(x, h_2, s_2))))$ where $\circ$ denotes element-wise multiplication.

Table 3. Basic statistics of the datasets used for evaluation

| Datasets | # training | # testing | # classes |
|---|---|---|---|
| CUB [35] | 5,994 | 5,794 | 200 |
| Aircraft [25] | 6,667 | 3,333 | 100 |
| Cars [17] | 8,144 | 8,041 | 196 |

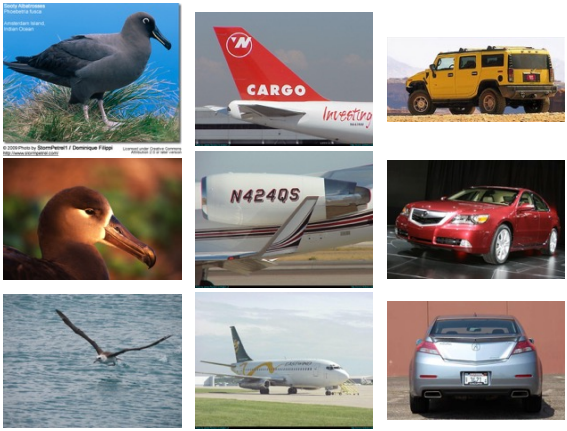

Figure 2. Sample images from the fine-grained classification datasets. From left to right, each column corresponds to CUB, Aircraft and Cars, respectively.

## 4.1. Experimental setup

We evaluated MoNet on three widely used fine-grained classification datasets. Different from general object recognition tasks, fine-grained classification usually tries to distinguish objects at the sub-category level, such as different makes of cars or different species of birds. The main challenge of this task is the relatively large inter-class and relatively small intra-class variations.

In all experiments, the 13 convolutional layers of VGG-16 [30] are used as the local feature extractor, and their outputs are used as local appearance representations. These 13 convolution layers are trained with ImageNet [7] and fine tuned in our experiments with three fine-grained classification datasets.

### 4.1.1 Datasets

**Caltech-UCSD birds (CUB) [35]** contains 200 species, mostly north-American, of birds. Being consistent with other works, we also use the 2011 extension with doubled number samples.

**FGVC-Aircraft Benchmark (Aircraft) [25]** is a benchmark fine-grained classification dataset with different aircrafts with various models and manufacturers.

**Stanford cars (Cars) [17]** contains images of different classes of cars at the level of make, model and year.

We use the provided train/test splits for all three datasets. Detailed information is given in Table 3 and Fig. 2 shows sample images.

### 4.1.2 Different pooling methods

**Bilinear pooling (BCNN):** The VGG-16 based BCNN [22] is utilized as the baseline pooling method, which applies the tensor product on the output of the conv$_{5\_3}$ layer with ReLU activation. The dimension of the final representation is $512 \times 512 \approx 262K$ and the number of the linear classifier parameters is $k \times 262K$, where $k$ is the number of classes. To be fair, the latest results from the authors' project page [1] are compared.

**Improved bilinear pooling (iBCNN):** Lin et al. [21] improved the original BCNN by adding the matrix power normalization after the bilinear pooling layer. We compare the results reported in [21] with VGG-16 as the back-bone network.

**Global Gaussian distribution embedding ($G^2$DeNet):** Instead of fully bilinear pooling, $G^2$DeNet pools the local features with a global Gaussian distribution embedding method, followed by a matrix square-root normalization.

Since it includes the first order moment information, the dimension of the final feature is slightly greater than BCNN and iBCNN. The experiment results with "w/o BBox" configuration in [36] are compared in this paper.

**Proposed moment embedding network (MoNet) and its variants:** We implemented the proposed MoNet architecture with structure as shown in Fig. 1 and fine-tuned the whole network in an end-to-end fashion. When using bilinear pooling, the feature dimensionality, computation and memory complexity are the same as $G^2$DeNet. To evaluate the effectiveness of the proposed layers HM and Ssqrt, we also tested MoNet variants. Depending on the left-out layer, we can have four different variants in total. Modifiers '2' and 'U' indicate that only 2nd order moments are incorporated during feature embedding, and that no normalization was used, respectively.

**Tensor Sketch compact pooling (TS):** When building the network with compact pooling, the TS layer [9] was added after the sub-matrix square-root layer. The projection dimension $D$ was selected empirically for MoNet and its variants.

### 4.1.3 Implementation details

Using a large enough number of samples is important to estimate stable and meaningful statistical moment information. The input images are resized to $448 \times 448$ in all the experiments, which produces a $28 \times 28 \times 512$ local feature matrix after conv$_{5\_3}$ for each image. Following common practice [36, 5], we first resize the image with a fixed aspect-ratio, such as the shorter edge reaches to 448 and then utilized a center crop to resize the image to $448 \times 448$. During training, random horizontal flipping was applied as data augmentation. Different from [21] with VGG-M, no augmentation is applied during testing.

To avoid rank deficiency, the singular value threshold $\sigma$ was set to $10^{-5}$ for both forward and backward propagation, which results in $10^{-10}$ for the singular value threshold of the tensor product matrix. The projected dimension in Tensor Sketch was fixed to $D = 10^4$, which satisfies $C < D \ll C^2$. For a smooth and stable training, we applied gradient clipping[26] to chop all gradients in the range $[-1, 1]$.

As suggested by [21, 36], all pooling methods were followed by an element-wise sign kept squre-root $\mathbf{y}_s = sign(\mathbf{y})\sqrt{\mathbf{y}}$ and $\ell_2$ normalization $\mathbf{y}_n = \mathbf{y}_s/||\mathbf{y}_s||$. For the sake of a smooth training, the element-wise square-root is also applied on local appearance features [36].

The weights of the VGG-16 convolutional layers are pretrained on ImageNet classification dataset. We first warm-started by fine-tuning the last linear classifier for 300 epochs. Then, we fine-tuned the whole network end-to-end with the learning rate as 0.001 and batch size as 16. The momentum was set to 0.9 and the weight decay was set to 0.0005. Most experiments converged to a local optimum after 50 epochs.

The proposed MoNet was implemented with MatConvNet [34] and Matlab 2017a[3]. Because of the numerical instability of SVDs, as suggested by Ionescu *et al.* [13], the sub-matrix square-root layer was implemented on CPU with double precision. The whole network was fine-tuned on a Ubuntu PC with 64GB RAM and Nvidia GTX 1080 Ti.

### 4.2. Experimental results

In Table 4 and Table 5, the classification accuracy for each network is presented in a row. Bilinear and TS denote fully bilinear pooling and tensor sketch compact pooling, respectively.

**Comparison with different variants:** The variants MoNet-2U, MoNet-2, and MoNet, when using bilinear pooling, are mathematically equivalent to BCNN, iBCNN, and $G^2$DeNet, respectively. Aligned with the observation in [21, 36], we also see a consistent performance gain for both MoNet and MoNet-2 by adding the normalization sub-matrix square root (SSqrt) layer. Specifically, MoNet-2 outperforms MoNet-2U by 0.6% to 1% with bilinear pooling and 0.6% to 0.8% with TS. Whereas MoNet outperforms MoNet-U by 3% to 4.9% with bilinear pooling and 0.8% to 0.9% with TS. This layer is more effective on MoNet than on MoNet-2. The reason for this, is that mixing different order moments may make the embedded feature numerically unstable but a good normalization helps overcome this issue. By adding the HM layer to incorporate 1st order moment information, MoNet can achieve better results consistently when compared to MoNet-2, in all datasets with both bilinear and compact pooling. Note that MoNet-U performs worse than MoNet-2U, which actually illustrates the merit of a proper normalization.

**Comparison with different architectures:** Consistent with [21], matrix normalization improves the performance by 1-2% on all three datasets. Our equivalent MoNet-2 achieves slightly better classification accuracy (0.2%) on CUB dataset but performs worse on Airplane and Car datasets when compared with iBCNN. We believe that this is due to the different approaches used to deal with rank deficiency. In our implementation, the singular value is hard thresholded as shown in Eq. 9, while iBCNN [21] dealt with the rank deficiency by adding 1 to all the singular values, which is a relatively very small number compared to the maximum singular value ($10^6$). By adding the 1st order moment information, $G^2$DeNet outperforms iBCNN by around 1%, on all three datasets. By re-writing the Gaussian embedding with tensor product of the homogeneous padded local features, our proposed MoNet can obtain similar or slightly better classification accuracy when compar-ing against G$^2$DeNet. Specifically, the classification accuracy of MoNet is 0.3% higher on Airplane dataset, but 0.7% lower on both CUB and Car datasets.

---
[3]Code is available at https://github.com/NEU-Gou/MoNet

Table 4. Experimental results for MoNet variants. Modifiers '2' and 'U' indicate that only 2nd order moments are incorporated during feature embedding, and that no normalization was used, respectively. The abbreviations for proposed layers are denoted as: **SSqrt**: sub-matrix square root; **HM**: Homogeneous mapping. The best result in each column is marked in red.

| New name | Missing layers | CUB | | Airplane | | Cars | |
| --- | --- | --- | --- | --- | --- | --- | --- |
| | | Bilinear | TS | Bilinear | TS | Bilinear | TS |
| MoNet-2U | HM, Ssqrt | 85.0 | 85.0 | 86.1 | 86.1 | 89.6 | 89.5 |
| MoNet-2 | HM | 86.0 | **85.7** | 86.7 | 86.7 | 90.5 | 90.3 |
| MoNet-U | Ssqrt | 82.8 | 84.8 | 84.4 | 87.2 | 88.8 | 90.0 |
| MoNet | - | **86.4** | **85.7** | **89.3** | **88.1** | **91.8** | **90.8** |

Table 5. Experimental results on fine-grained classification. Bilinear and TS represent fully bilinear pooling and Tensor Sketch compact pooling respectively. The best performance in each column is marked in red.

| | | CUB | | Airplane | | Car | |
| --- | --- | --- | --- | --- | --- | --- | --- |
| | | Bilinear | TS | Bilinear | TS | Bilinear | TS |
| | BCNN [22, 9] | 84.0 | 84.0 | 86.9 | 87.2 | 90.6 | 90.2 |
| | MoNet-2U | 85.0 | 85.0 | 86.1 | 86.1 | 89.6 | 89.5 |
| | iBCNN [21] | 85.8 | - | 88.5 | - | 92.1 | - |
| | MoNet-2 | 86.0 | **85.7** | 86.7 | 86.7 | 90.5 | 90.3 |
| | G$^2$DeNet [36] | **87.1** | - | 89.0 | - | **92.5** | - |
| | MoNet | 86.4 | **85.7** | **89.3** | **88.1** | 91.8 | **90.8** |
| Other higher | KP [5] | - | **86.2** | - | 86.9 | - | **92.6** |
| order methods | HOHC [2] | 85.3 | | 88.3 | | 91.7 | |
| State-of-the-art | MA-CNN [38] | 86.5 | | 89.9 | | 92.8 | |

**Comparison with fully bilinear pooling and compact pooling:** As shown in [9], compact pooling can achieve similar performance compared to BCNN, but with only 4% of the dimensionality. We also see a similar trend in MoNet-2U and MoNet-2. The classification accuracy difference between the bilinear pooling and compact pooling version is less than 0.3% on all three datasets. However, the performance gaps are relatively greater when we compare the different pooling schemes on MoNet. Bilinear pooling improve the classification accuracy by 0.7%, 1.2% and 1% than compact pooling on CUB, Airplane and Car datasets, respectively. However, with compact pooling, the dimensionality of the final representation is 96% smaller. Although the final fully bilinear pooled representation dimensions of MoNet-2 and MoNet are roughly the same, MoNet utilizes more different order moments, which requires more count sketch projections to approximate it. Thus, when fixing $D = 10,000$ for both MoNet-2 and MoNet, the performance of MoNet with compact pooling degraded. However, MoNet TS still out-performs MoNet-2 TS by 1.4% and 0.5% on the Airplane and Car datasets, respectively.

**Comparison with other methods:** [5] and [2] are two other recent works that also take into account higher order statistic information. Cui *et al*. [5] applied Tensor Sketch repetitively to approximate up to 4th order explicit polynomial kernel space in a compact way. They obtained better results for CUB and Car datasets compared against other compact pooling results, but notably worse (1.2%) on the Airplane dataset. This may be due to two factors. First, directly utilizing higher order moments without proper normalization leads to numerically instabilities. Second, approximating higher order moments with limited number of samples is essentially an ill-posed problem. Cai *et al*. [2] only utilize higher order self-product terms but not the interaction terms, which leads to worse performance in all three datasets. Finally, the state-of-the-art MA-CNN [38] achieves slightly better results on Airplane and Car datasets.

## 5. Conclusion

Bilinear pooling, as a recently proposed 2nd order moment pooling method, has been shown effective in several vision tasks. iBCNN [21] improves the performance with matrix square-root normalization and G$^2$DeNet [36] extends it by adding 1st order moment information. One key limitation of these approaches is the high dimension of the final representation. To resolve this, compact pooling methods have been proposed to approximate the bilinear pooling. However, two factors make using compact pooling on iBCNN and G$^2$DeNet non-trivial. Firstly, the Gaussian embedding formation entangles the bilinear pooling. Secondly, matrix normalization needs to be applied on the bilinear pooled matrix. In this paper, we reformulated the Gaussian embedding using the empirical moment matrix and decoupled the bilinear pooling step out. With the help of a novel sub-matrix square-root layer, our proposed network MoNet can take advantages of different order moments, matrix normalization as well as compact pooling. Experiments on three widely used fine-grained classification datasets demonstrate that MoNet can achieve similar or better performance when comparing with G$^2$DeNet and retain comparable results with only 4% of the feature dimensions.

## References

[1] Bilinear CNNs project. http://vis-www.cs.umass.edu/bcnn/. Accessed: 2018-03-27.

[2] S. Cai, W. Zuo, and L. Zhang. Higher-order integration of hierarchical convolutional activations for fine-grained visual categorization. In *Proceedings of the IEEE Conference on Computer Vision and Pattern Recognition*, pages 511–520, 2017.

[3] J. Carreira, R. Caseiro, J. Batista, and C. Sminchisescu. Semantic segmentation with second-order pooling. *Computer Vision–ECCV 2012*, pages 430–443, 2012.

[4] M. Cimpoi, S. Maji, and A. Vedaldi. Deep filter banks for texture recognition and segmentation. In *Proceedings of the IEEE Conference on Computer Vision and Pattern Recognition*, pages 3828–3836, 2015.

[5] Y. Cui, F. Zhou, J. Wang, X. Liu, Y. Lin, and S. Belongie. Kernel pooling for convolutional neural networks. In *Computer Vision and Pattern Recognition (CVPR)*, 2017.

[6] N. Dalal and B. Triggs. Histograms of oriented gradients for human detection. In *Computer Vision and Pattern Recognition, 2005. CVPR 2005. IEEE Computer Society Conference on*, volume 1, pages 886–893. IEEE, 2005.

[7] J. Deng, W. Dong, R. Socher, L.-J. Li, K. Li, and L. Fei-Fei. Imagenet: A large-scale hierarchical image database. In *Computer Vision and Pattern Recognition, 2009. CVPR 2009. IEEE Conference on*, pages 248–255. IEEE, 2009.

[8] A. Fukui, D. H. Park, D. Yang, A. Rohrbach, T. Darrell, and M. Rohrbach. Multimodal compact bilinear pooling for visual question answering and visual grounding. *arXiv preprint arXiv:1606.01847*, 2016.

[9] Y. Gao, O. Beijbom, N. Zhang, and T. Darrell. Compact bilinear pooling. In *Proceedings of the IEEE Conference on Computer Vision and Pattern Recognition*, pages 317–326, 2016.

[10] L. A. Gatys, A. S. Ecker, and M. Bethge. A neural algorithm of artistic style. *arXiv preprint arXiv:1508.06576*, 2015.

[11] Y. Gong, L. Wang, R. Guo, and S. Lazebnik. Multi-scale orderless pooling of deep convolutional activation features. In *European conference on computer vision*, pages 392–407. Springer, 2014.

[12] M. Gou, O. Camps, and M. Sznaier. mom: Mean of moments feature for person re-identification. In *The IEEE International Conference on Computer Vision Workshop (ICCVW)*, 2017.

[13] C. Ionescu, O. Vantzos, and C. Sminchisescu. Matrix backpropagation for deep networks with structured layers. In *Proceedings of the IEEE International Conference on Computer Vision*, pages 2965–2973, 2015.

[14] H. Jégou, M. Douze, C. Schmid, and P. Pérez. Aggregating local descriptors into a compact image representation. In *Computer Vision and Pattern Recognition (CVPR), 2010 IEEE Conference on*, pages 3304–3311. IEEE, 2010.

[15] P. Kar and H. Karnick. Random feature maps for dot product kernels. In *International conference on artificial intelligence and statistics*, pages 583–591, 2012.

[16] S. Kong and C. Fowlkes. Low-rank bilinear pooling for fine-grained classification. In *The IEEE Conference on Computer Vision and Pattern Recognition (CVPR)*, July 2017.

[17] J. Krause, M. Stark, J. Deng, and L. Fei-Fei. 3d object representations for fine-grained categorization. In *Proceedings of the IEEE International Conference on Computer Vision Workshops*, pages 554–561, 2013.

[18] A. Krizhevsky, I. Sutskever, and G. E. Hinton. Imagenet classification with deep convolutional neural networks. In *Advances in neural information processing systems*, pages 1097–1105, 2012.

[19] J.-B. Lasserre and E. Pauwels. Sorting out typicality with the inverse moment matrix sos polynomial. In *Neural Information Processing Systems (NIPS 2016)*, 2016.

[20] P. Li, J. Xie, Q. Wang, and W. Zuo. Is second-order information helpful for large-scale visual recognition? In *The IEEE International Conference on Computer Vision (ICCV)*, Oct 2017.

[21] T.-Y. Lin and S. Maji. Improved bilinear pooling with cnns. In *BMVC*, 2017.

[22] T.-Y. Lin, A. RoyChowdhury, and S. Maji. Bilinear cnn models for fine-grained visual recognition. In *The IEEE International Conference on Computer Vision (ICCV)*, December 2015.

[23] M. Lovrić, M. Min-Oo, and E. A. Ruh. Multivariate normal distributions parametrized as a riemannian symmetric space. *Journal of Multivariate Analysis*, 74(1):36–48, 2000.

[24] D. G. Lowe. Object recognition from local scale-invariant features. In *Computer vision, 1999. The proceedings of the seventh IEEE international conference on*, volume 2, pages 1150–1157. Ieee, 1999.

[25] S. Maji, J. Kannala, E. Rahtu, M. Blaschko, and A. Vedaldi. Fine-grained visual classification of aircraft. Technical report, 2013.

[26] R. Pascanu, T. Mikolov, and Y. Bengio. On the difficulty of training recurrent neural networks. In *International Conference on Machine Learning*, pages 1310–1318, 2013.

[27] F. Perronnin, J. Sánchez, and T. Mensink. Improving the fisher kernel for large-scale image classification. *Computer Vision–ECCV 2010*, pages 143–156, 2010.

[28] N. Pham and R. Pagh. Fast and scalable polynomial kernels via explicit feature maps. In *Proceedings of the 19th ACM SIGKDD international conference on Knowledge discovery and data mining*, pages 239–247. ACM, 2013.

[29] A. RoyChowdhury, T.-Y. Lin, S. Maji, and E. Learned-Miller. Face identification with bilinear cnns. *arXiv preprint arXiv: 1506.01342*, 2015.

[30] K. Simonyan and A. Zisserman. Very deep convolutional networks for large-scale image recognition. *arXiv preprint arXiv:1409.1556*, 2014.

[31] M. Sznaier and O. Camps. Sos-rsc: A sum-of-squares polynomial approach to robustifying subspace clustering algorithms. In *The IEEE Conference on Computer Vision and Pattern Recognition (CVPR)*, June 2018.

[32] J. B. Tenenbaum and W. T. Freeman. Separating style and content. In *Advances in neural information processing systems*, pages 662–668, 1997.

[33] O. Tuzel, F. Porikli, and P. Meer. Pedestrian detection via classification on riemannian manifolds. *IEEE transactions on pattern analysis and machine intelligence*, 30(10):1713–1727, 2008.

[34] A. Vedaldi and K. Lenc. Matconvnet – convolutional neural networks for matlab. In *Proceeding of the ACM Int. Conf. on Multimedia*, 2015.

[35] C. Wah, S. Branson, P. Welinder, P. Perona, and S. Belongie. The Caltech-UCSD Birds-200-2011 Dataset. Technical report, 2011.

[36] Q. Wang, P. Li, and L. Zhang. G2denet: Global gaussian distribution embedding network and its application to visual recognition. In *The IEEE Conference on Computer Vision and Pattern Recognition (CVPR)*, July 2017.

[37] Z. Yu, J. Yu, J. Fan, and D. Tao. Multi-modal factorized bilinear pooling with co-attention learning for visual question answering. *arXiv preprint arXiv:1708.01471*, 2017.

[38] H. Zheng, J. Fu, T. Mei, and J. Luo. Learning multi-attention convolutional neural network for fine-grained image recognition. In *Int. Conf. on Computer Vision*, 2017.